\pdfoutput=1

\documentclass[11pt]{article}

\usepackage{acl}

\usepackage{times}
\usepackage{latexsym}
\usepackage{booktabs}
\usepackage{amsfonts}

\usepackage[T1]{fontenc}

\usepackage[utf8]{inputenc}

\usepackage{microtype}

\usepackage[colorinlistoftodos,textsize=tiny]{todonotes}

\title{Mintaka: A Complex, Natural, and Multilingual Dataset for End-to-End Question Answering}

\author{Priyanka Sen \\
  Amazon Alexa AI \\
  Cambridge, UK \\
  \texttt{sepriyan@amazon.com} \\\And
  Alham Fikri Aji \\
  Amazon Alexa AI \\
  Cambridge, UK \\
  \texttt{afaji@amazon.com} \\\And
  Amir Saffari \\
  Amazon Alexa AI \\
  Cambridge, UK \\
  \texttt{amsafari@amazon.com} \\}

\begin{document}
\maketitle
\begin{abstract}
We introduce \textsc{Mintaka}, a complex, natural, and multilingual dataset designed for experimenting with end-to-end question-answering models. Mintaka is composed of 20,000 question-answer pairs collected in English, annotated with Wikidata entities, and translated into Arabic, French, German, Hindi, Italian, Japanese, Portuguese, and Spanish for a total of 180,000 samples. Mintaka includes 8 types of complex questions, including superlative, intersection, and multi-hop questions, which were naturally elicited from crowd workers. We run baselines over Mintaka, the best of which achieves 38\% hits@1 in English and 31\% hits@1 multilingually, showing that existing models have room for improvement. We release Mintaka at \url{https://github.com/amazon-research/mintaka}.
\end{abstract}

\section{Introduction}

Question answering (QA) is the task of learning to predict answers to questions. Approaches to question answering include knowledge graph (KG) based methods, which use structured data to find the correct answer \cite{miller-etal-2016-key,saxena-etal-2020-improving}; machine reading comprehension methods, which extract answers from input documents \cite{rajpurkar2016squad,kwiatkowski2019natural}; open domain methods, which learn to retrieve relevant documents and extract or generate answers \cite{zhu2021retrieving}, and closed-book methods, which use knowledge implicitly stored in model parameters to answer questions \cite{2020t5cqba}. 

With state-of-the-art techniques, QA models can achieve high performance on simple questions \cite{shi2020kqa,shi-etal-2021-transfernet} that require a single fact lookup in either a knowledge graph or a text document (e.g., "\textit{Where was Natalie Portman born?}"). However not all questions are simple in real-world applications. We define \textit{complex questions} \cite{ijcai2021-611} as questions that require an operation beyond a single fact lookup, such as multi-hop, comparative, or set intersection questions. For example, "\textit{What movie had a higher budget, Titanic or Men in Black?}" requires looking up the budget of two movies, comparing the values, and selecting the movie with the higher budget. Handling more complex questions remains an open problem.

One of the challenges in measuring and improving QA performance on complex questions is a lack of datasets. While several QA datasets exist, they have shortcomings of being either large but simple, such as SimpleQuestions \cite{bordes2015large}, or complex but small, such as ComplexQuestions \cite{bao2016constraint} or QALD \cite{usbeck2018}. Recently, several large and complex datasets have been released, including KQA Pro \cite{shi2020kqa} and GrailQA \cite{gu2021beyond}. These datasets use automatically generated questions followed by human paraphrasing, which can result in less natural questions, such as "\textit{Is the WOEID of Tuscaloosa 14605?}" (KQA Pro) or "\textit{1520.0 is the minimum width for which size rail gauge}?" (GrailQA). This can lead to a mismatch between training data and real-world use cases of QA models.

\begin{table*}[htb!]
\begin{center}
\begin{tabular}{l c c c c c}
\toprule
Dataset & Samples & Text or KG & Complex & Natural & Languages\\
\midrule
SQuAD & 150K & Wikipedia & $\times$ & $\times$ & 1\\
XQuAD & 2K & Wikipedia & $\times$ & $\times$ & 11\\
Natural Questions & 300K & Wikipedia & $\times$ & \checkmark & 1\\
HotpotQA & 100K & Wikipedia & \checkmark & $\times$ & 1 \\
DROP & 100K & Wikipedia & \checkmark & $\times$ & 1\\
WebQuestionsSP & 5K & FreeBase & $\times$ & \checkmark & 1\\
ComplexQuestions & 2K & FreeBase & \checkmark & \checkmark & 1\\
ComplexWebQuestions & 35K & FreeBase & \checkmark & $\times$ & 1\\
LC-QuAD 2.0 & 30K & Wikidata, DBPedia & \checkmark & $\times$ & 1\\
GrailQA & 64K & Wikidata & \checkmark & $\times$ & 1\\
KQA Pro & 120K & Wikidata & \checkmark & $\times$ & 1\\
QALD & 400 & DBPedia & \checkmark & \checkmark & 11\\
\textbf{Mintaka (ours)} & 20K & Wikidata & \checkmark & \checkmark & 9\\
\bottomrule
\end{tabular}
\end{center}
\caption{Comparison of Mintaka to existing QA datasets}
\label{tab:dataset-comparison}
\end{table*}

In order to address these gaps, we release \textsc{Mintaka}, a large, complex, naturally-elicited, and multilingual question answering dataset. Mintaka contains 20,000 question-answer pairs elicited in English from crowd workers. We link Mintaka to a knowledge graph by asking crowd workers to annotate the question and answer text with Wikidata IDs. Professional translators translated the 20,000 English questions into Arabic, French, German, Hindi, Italian, Japanese, Portuguese, and Spanish, creating a total dataset size of 180,000 questions. 

In this paper, we present an overview of Mintaka in §\ref{mintaka}, explain how we built Mintaka in §\ref{dataset_collection}, provide a statistical analysis of the dataset in §\ref{dataset_analysis}, including a demographic analysis of our crowd workers in §\ref{demographics}. Finally in §\ref{baselines}, we present results of existing baseline models on Mintaka, the best of which scores 38\% hits@1. These results show that existing models have room for improvement. 

We publicly release the Mintaka dataset with our randomly split train (14,000 samples), dev (2,000 samples), and test (4,000 samples) sets at \url{https://github.com/amazon-research/mintaka}.

\section{Related Works}

Question answering has no shortage of datasets. Datasets for question-answering with reading comprehension, such as SQuAD \cite{rajpurkar2016squad} or Natural Questions \cite{kwiatkowski2019natural} are often large, and some are even multilingual, such as XQuAD \cite{Artetxe:etal:2019}, MLQA \cite{lewis2019mlqa}, or TyDi QA \cite{clark2020tydi}. These datasets, however, are not explicitly built to be complex, and the answer is usually found in a single passage of text. 

HotpotQA \cite{yang-etal-2018-hotpotqa} and MuSiQue \cite{trivedi2022musique} add complexity to reading comprehension by introducing multi-hop questions where the answer requires reasoning over two documents, but neither of these datasets naturally elicit their questions. HotpotQA pre-selects two Wikipedia passages and asks workers to write questions using both passages, and MuSiQue composes multi-hop questions from existing single hop questions. DROP \cite{dua-etal-2019-drop} is another complex reading comprehension dataset, including complex operations such as addition, counting, and sorting. Again, DROP asks crowd workers to write questions about a selected Wikipedia passage. DROP additionally introduces a constraint where workers need to write questions that can't be solved by an existing model.

Within knowledge graph-based question answering (KGQA), WebQuestionsSP \cite{berant2013semantic,yih2016value} and ComplexQuestions \cite{bao2016constraint} are more natural QA datasets. Both collected real user questions using search query logs or the Google Suggest API. The answers were annotated manually using FreeBase as a knowledge graph. WebQuestionsSP contains mostly simple questions, but ComplexQuestions is more complex, including multi-hop questions, temporal constraints, and aggregations. The main drawback of these datasets is size. WebQuestionsSP contains 5K QA pairs, while ComplexQuestions contains only 2K. 

ComplexWebQuestions \cite{talmor2018web} is a dataset based on WebQuestionsSP, which increases the size to 35K QA pairs and introduces more complex operations, including multi-hop, comparatives, and superlatives. However ComplexWebQuestions loses some naturalness, as the dataset is built by automatically generating queries and questions, and then asking crowd workers to paraphrase the generated questions.

Recently, several larger-scale complex KGQA datasets have been released. LC-QuAD 2.0 \cite{dubey2017lc2} includes 30K questions, including multi-hop questions, and uses the more up-to-date Wikidata and DBpedia knowledge graphs. GrailQA \cite{gu2021beyond} is even larger at 64K questions based on FreeBase with complex questions, including multi-hop, count, and comparatives. KQA Pro \cite{shi2020kqa} is even larger still with 120K questions based on Wikidata and with complex questions, including intersection and superlatives. All these datasets make the trade-off of scale over naturalness. To collect question-answer pairs, the authors generate queries from a knowledge graph, generate questions based on the queries, and then ask crowd workers to paraphrase the questions. 

Finally, most datasets are only in English. Multilingual and complex datasets are rare. QALD 2018 \cite{usbeck2018} is one multilingual and complex dataset including 11 languages and complex operations such as counts and comparatives, however includes only 400 questions. 

By building Mintaka, we hope to address an important gap in existing datasets. Mintaka question-answer pairs are both complex and naturally-elicited from crowd workers with no restrictions on what facts or articles the questions can be about. We also translate Mintaka into 8 languages, making it one of the first large-scale complex and multilingual question answering datasets. A comparison of Mintaka to existing datasets can be seen in Table \ref{tab:dataset-comparison}.

\section{Mintaka} \label{mintaka}

Mintaka is a complex question answering dataset of 20,000 questions collected in English and translated into 8 languages, for a total of 180,000 questions. Mintaka contains question-answer pairs written by crowd workers and annotated with Wikidata entities in both the question and answer. 

We collected questions in eight topics, which were chosen for being broadly appealing and suitable for writing complex questions: \textsc{Movies, Music, Sports, Books, Geography, Politics, Video Games,} and \textsc{History}. Since we want Mintaka to be a complex question answering dataset, we explicitly collected questions in the following complexity types: (Note: All examples below are from the Mintaka dataset.)

\begin{itemize}
\item \textsc{Count}: questions where the answer requires counting. For example, Q: How many astronauts have been elected to Congress? A: 4
\item \textsc{Comparative}: questions that compare two objects on a given attribute (e.g., age, height). For example, Q: Is Mont Blanc taller than Mount Rainier? A: Yes
\item \textsc{Superlative}: questions about the maximums or minimums of a given attribute. For example, Q: Who was the youngest tribute in the Hunger Games? A: Rue
\item \textsc{Ordinal}: questions based on an object's position in an ordered list. For example, Q: Who was the last Ptolemaic ruler of Egypt? A: Cleopatra
\item \textsc{Multi-hop}: questions that require 2 or more steps (multiple \textit{hops}) to answer. For example, Q: Who was the quarterback of the team that won Super Bowl 50? A: Peyton Manning
\item \textsc{Intersection}: questions that have two or more conditions that the answer must fulfill. For example, Q: Which movie was directed by Denis Villeneuve and stars Timothee Chalamet? A: Dune
\item \textsc{Difference}: questions with a condition that contains a negation. For example, Q: Which Mario Kart game did Yoshi not appear in? A: Mario Kart Live: Home Circuit
\item \textsc{Yes/No}: questions where the answer is Yes or No. For example, Q: Has Lady Gaga ever made a song with Ariana Grande? A: Yes.
\item \textsc{Generic:} questions where the worker was only given the topic and no constraints of complexity. These tend to be simpler fact lookups, such as Q: Where was Michael Phelps born? A: Baltimore, Maryland 
\end{itemize}

For each of the 8 topics, we collected 250 questions per complexity type and 500 generic questions, for a total of 2,500 questions per topic.

We also collected translations of the 20,000 English questions in 8 languages using professional translators. Since all questions were collected in English from U.S. workers, the questions may have a U.S. bias in terms of the entities (for example, U.S. politicians or books written in English). This is a choice we make since it allows us to create a fully parallel dataset where models can be easily compared across languages. This choice was also made in previous QA datasets \cite{usbeck2018,Artetxe:etal:2019,lewis2019mlqa}.

\section{Dataset Collection} \label{dataset_collection}

To build our dataset, we used Amazon Mechanical Turk (MTurk) in three different tasks. All of our MTurk workers were located in the United States, and to ensure high quality, we required workers have an approval rating of 98\% and at least 5,000 approved tasks. Each of our tasks are explained in the sections below, and examples of the interfaces can be seen in Appendix \ref{appendix:a}.

\subsection{Question Elicitation}

The first task was to elicit complex questions. To do this, we created tasks for each topic/complexity pair (e.g., Superlative Movie questions, Ordinal Sports questions, etc.). In each task, a worker was asked to write 5 questions and answers about the topic using the given complexity type. The questions and answers were written in free text fields. We had no restrictions on what sources workers could use to write their questions, so workers were not limited to writing questions based on a given article or facts. Workers were given explanations of the complexity type and examples in the instructions. The topics were left general, so within History, workers could write about Ancient Egypt as well as World War II. 

For Count and Superlative answers, we additionally asked workers to provide a numerical value as part of the answer. For example, in Count questions, workers would both provide the answer as a number (e.g., 3) as well as the entities that make up that answer (e.g., Best Picture, Best Adapted Screenplay, and Best Film Editing). In Superlative questions, workers provided the answer (e.g., Missouri River) as well as the numerical value that makes the entity the maximum or minimum (e.g., 2,341 miles). Additionally in Count questions, if a question had multiple answers, we asked workers to list a minimum of five. For example, for the question "\textit{How many cities have hosted a Summer Olympics?}", a worker could give the numerical answer 23 but provide only five of the cities. For this reason, answers to questions with more than five entities are not guaranteed to be complete but instead provide a sample of the correct answer.

We paid \$1.25 per task to write five questions. Workers were limited to completing one task per topic-complexity pair. After collection, we also surveyed the MTurk workers who completed our Question Elicitation task about their demographics. The results of this survey are discussed in §\ref{demographics}.

\subsection{Answer Entity Linking}

Answers were collected in the previous task in natural language. In order to link the answers to a knowledge graph, we built an Answer Entity Linking task. We chose to link the answers to Wikidata, since it is a large and up-to-date public knowledge graph. Although we link to Wikidata, we don't guarantee that every question can be answered by Wikidata at the time of writing. It is possible that there are missing or incomplete facts that would prevent a KGQA system from reaching the answer entity in Wikidata given the question.

In this task, workers were shown a question-answer pair and asked to 1) highlight the entities in the answer, and 2) search for the entities on Wikidata and provide the correct URLs. We built a UI for MTurk workers where they could easily highlight entities and the highlighted entities would automatically generate links to search Wikidata. 

Each answer was annotated by two MTurk workers. For agreement, we required two workers to identify the same entities and the same Wikidata URLs for all entities. If there was disagreement, we sent the question-answer pair to a third annotator. Question-answer pairs where the answer was a number or \textit{yes} or \textit{no} were excluded from answer entity linking. Overall, we annotated 20,996 answer entities and achieved 82\% agreement after two annotators and 97\% agreement after three annotators. The remaining 3\% were verified by the authors.

We paid a base rate of \$0.10 per task, which consisted of a single question-answer pair. If the answer had multiple entities, we paid a \$0.05 bonus for every additional entity identified that was agreed upon by another annotator. 

\begin{table*}[t!]
\begin{minipage}{.50\linewidth}
\centering
\begin{tabular}{l r}
\toprule
\textbf{Question Length} &\\
English & 10.2\\
Arabic & 9.9\\
German & 9.6\\
Spanish & 10.8\\
French & 12.4\\
Hindi & 11.0\\
Italian & 10.6\\
Japanese (in characters) & 29.6\\
Portuguese & 10.3\\
\midrule
\textbf{Entities} &\\
Entities per Question & 1.8\\
Entities per Answer & 1.3\\
Unique Question Entities & 7,289\\
Unique Answer Entities & 8,605\\
Unique Entities & 13,214\\
Question to Answer Entity &\\
\hspace{10pt} within one hop & 62\%\\
\hspace{10pt} within two hops & 97\%\\
\midrule
\textbf{Answer Types} & \\
Entity & 0.72\\
Boolean & 0.14\\
Numerical & 0.07\\
Date & 0.06\\
String & 0.001\\
\bottomrule
\end{tabular}
\caption{Statistics about the Mintaka dataset}
\label{tab:stats}
\end{minipage}
\begin{minipage}{.5\linewidth}

\begin{tabular}{l}
\toprule
\textbf{Q:} Which Studio Ghibli (Q182950) movie scored \\
the lowest on Rotten Tomatoes (Q105584)?\\
\textbf{A:} Earwig and the Witch (Q96031360)\\
\midrule
\textbf{Q:} Which revolution lasted longer, the French\\
(Q6534) or the American (Q192769)?\\
\textbf{A:} American Revolution (Q192769)\\
\midrule
\textbf{Q:} When Franklin D. Roosevelt (Q8007) was \\
first elected, how long had it been since someone \\
in his party (Q7278) won the presidential \\
election (Q47566)? \\
\textbf{A:} 16 years\\
\midrule
\textbf{Q:} Which member of the Red Hot Chili Peppers \\
(Q10708) appeared in Point Break (Q1146552)?\\
\textbf{A:} Anthony Kiedis (Q204751)\\
\midrule
\textbf{Q:} Which Mass Effect (Q953242) game does not \\
include Commander Shepard (Q3683919) as the \\
main character?\\
\textbf{A:} Mass Effect: Andromeda (Q20113552)\\
\bottomrule
\end{tabular}
\caption{Example question-answer pairs from Mintaka. Question and answer annotations are shown here in-line with Wikidata Q-codes.}
\label{tab:examples}
\end{minipage}
\end{table*}

\subsection{Question Entity Linking}

An end-to-end question answering model can be trained using the question and answer alone \cite{oliya-etal-2021-end}. However to better evaluate end-to-end methods and train models requiring entities, we also created an MTurk task to link entities in the question text. 

Linking entities in questions is more challenging than answers. While answer texts are often short and contain a clear entity (e.g., "Joe Biden"), question texts can contain multiple possible entities. In the question "\textit{Who is the president of the United States?}", a worker could select "United States", or "president" and "United States", or even "president of the United States". Since early test runs showed it would be difficult to get agreement on question entities, we modified the task so workers only verified a span and linked the entity in Wikidata.

To identify spans in questions, we used spaCy’s \cite{Honnibal2020} en\_core\_web\_trf model to identify named entities and noun chunks with capitalized words in the English sentences. We then gave workers the question with a predicted entity highlighted. Workers were shown one entity at a time and asked to first verify or modify the highlighted entity and then link to Wikidata. 

For question entities that were not seen before, we had the entity annotated by two annotators, followed by a third in case of no agreement. For some question entities, we were able to exact string match them against entities that were already annotated in the Answer Entity Linking task, for example United States → Q30. In cases where we had a match, the question entity was annotated by one annotator and only went to a second annotator if there was no agreement.

We annotated 12,819 new entities, for which we had 68\% agreement after two annotators and 78\% agreement after three annotators, and 15,075 seen entities, for which we had 80\% agreement after one annotator, and 98\% agreement after two annotators. The remaining entities were verified by the authors. We paid \$0.10 per entity. The spans of the question entities are only annotated in English, and so English questions in Mintaka come with both the entity ID and the span, while all translated questions have only the entity ID.

\subsection{Translations}

We translated the 20,000 questions in Mintaka to the following languages and locales: Arabic (Saudi Arabia), French (France), German (Germany), Hindi (India), Italian (Italy), Japanese (Japan), Portuguese (Brazil), and Spanish (Mexico). Translation is our only dataset collection step where we do not use MTurk. Early experiments with MTurk on translation tasks and editing automatic translations tasks had poor results and a lack of workers in some languages, such as Japanese. For this reason, we use professional translators.

\section{Dataset Analysis} \label{dataset_analysis}

\subsection{Dataset Statistics}

Statistics about the Mintaka dataset are shown in Table \ref{tab:stats}. Question length is based on white space splitting in all languages except Japanese, where the question length is in characters. In total, 13,232 unique Wikidata entities appear across all questions. The most common question entities are the United States (Q30; 1,495 questions), President of the United States (Q11696; 565 questions), and Super Bowl (Q32096, 345 questions). The most common answer entities are California (Q99, 102 answers), Alaska (Q797, 88 answers), and the United States (Q30, 80 answers). 

Mintaka was built with only questions and answers, so we do not know the correct query path. However we can calculate an upper bound of answerable questions using Wikidata by identifying the percentage of questions that have a path connecting the question entity to the answer entity. We find that there is a path connecting the question and answer entity 62\% of the time within 1 hop and 97\% of the time within 2 hops for questions linked to Wikidata entities.

A majority (72\%) of the questions in Mintaka can be answered using an entity. 14\% can be answered using a boolean, in yes/no or comparative questions. 7\% can be answered using a number, such as someone’s age. 6\% can be answered using a date, such as a date of birth. And finally, 0.1\% have answers in the form of a string, for example, someone’s nickname. Examples of QA pairs are in Table \ref{tab:examples} with more examples in Appendix \ref{appendix:c}.

\subsection{Naturalness Evaluation}

\begin{figure}[h!]
    \centering
    \includegraphics[width=7cm]{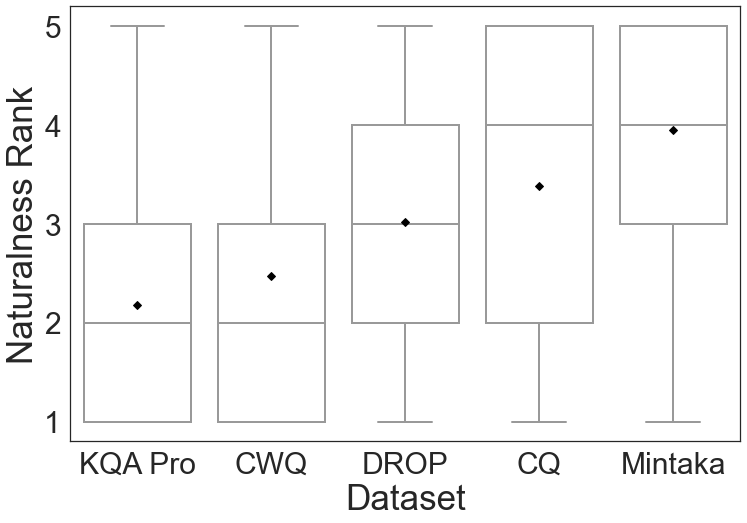}
    \caption{A box plot showing the quartile, median, and mean (black diamond) naturalness rank for each dataset from 1 (least natural) to 5 (most natural).}
\label{fig:naturalness}
\end{figure}

By naturally eliciting complex questions from MTurk workers, we aimed to collect questions that were closer to what users may ask in real-world settings. In order to evaluate how Mintaka compares to previous complex QA datasets, we ran a naturalness evaluation task on Mturk with four comparison datasets. We compared datasets that collected questions in different ways: KQA Pro automatically generated questions, ComplexWebQuestions (CWQ) automatically generated questions built off WebQuestions, DROP naturally elicited questions about a given Wikipedia passage, and ComplexQuestions (CQ) collected natural questions from user logs. We compared these datasets to Mintaka in a task where workers were shown 5 questions, one from each dataset, and asked to rank them from 1 (least natural) to 5 (most natural). We uniformly sampled 500 questions from each dataset and grouped them into quartiles by length for each dataset (i.e., the longest questions in Mintaka are grouped with the longest questions in other datasets). 

The results are in Figure \ref{fig:naturalness} and show that Mintaka is on average ranked higher in naturalness than all other datasets. We also find that Mintaka ranks significantly higher than the other datasets using a two-sample Kolmogorov-Smirnov test (with p-values < 0.001). This shows that Mintaka questions are perceived as more natural than automatically generated or passage-constrained questions. Although ComplexQuestions contains real user questions, the questions are collected from search logs and can be phrased ungrammatically (e.g., "\textit{when did miami dolphins win super bowl?}"), leading to a wider range of ranks. These results confirm that Mintaka is both a complex and natural dataset.

\subsection{Demographics of MTurk Workers} \label{demographics}

\begin{table}[ht!]
\begin{center}
\begin{tabular}{l r}
\toprule
\textsc{\textbf{Gender}} & \\
Male & 0.58 \\
Female & 0.42\\
\midrule
\textsc{\textbf{Age}} & \\
18-24 & 0.02\\
25-34 & 0.39\\
35-44 & 0.32\\
45-54 & 0.13\\
55-64 & 0.10\\
65+ & 0.04\\
\midrule
\textsc{\textbf{Ethnicity}} & \\
White & 0.73\\
Asian & 0.10\\
Black & 0.07\\
Hispanic & 0.06\\
Multiracial & 0.04\\
\midrule
\textsc{\textbf{Education}} & \\
High school & 0.16\\
Associate's & 0.23\\
Bachelor's & 0.50\\
Master's & 0.09\\
Doctoral & 0.02\\
\midrule
\textsc{\textbf{Employment}} & \\
Employed, full-time & 0.65\\
Self-employed & 0.17\\
Employed, part-time & 0.09\\
Not employed & 0.05\\
Retired & 0.02\\
Homemaker & 0.01\\
\midrule
\textsc{\textbf{Residential Area}} &\\
Urban & 0.33\\
Suburban & 0.52\\
Rural & 0.15\\
\midrule
\textsc{\textbf{U.S. Region}} &\\
Northeast & 0.21\\
South & 0.35\\
Midwest & 0.21\\
West & 0.23\\
\bottomrule
\end{tabular}
\end{center}
\caption{Results of the demographic survey of workers who completed the Question Elicitation task. Options that received less than 1\% of responses are not shown.}
\label{tab:demographics}
\end{table}

In total, 516 MTurk workers completed 3,503 Question Elicitation tasks to collect complex questions (some questions from tasks were removed as duplicates or under-sampled for a balanced dataset size). In order to better understand and measure who our dataset best represents, we invited all workers who completed a Question Elicitation task to participate in a demographic survey. We paid workers \$1.25 to complete the survey. We received 400 responses (78\% response rate). Worker IDs were only used to invite MTurk workers to take part in the survey. All demographic data is anonymous with no way to link the data back to the Worker IDs, and the data is only analyzed in aggregate.

Table \ref{tab:demographics} can be used as an indicator of who this dataset is best for modeling and for what populations it may be less representative. For example, we had more workers identify as male than female (58\% vs. 42\%). Only 2\% of workers were between the ages of 18-24 (and workers below 18 cannot register on MTurk), while 72\% of workers were between the ages of 25 and 44. We also had fewer workers who identified as Black (7\%) or Hispanic (6\%) than the U.S. Census \cite{us-census2021} estimates of the general population (13\% and 19\%, respectively), while seeing a slightly higher percentage of workers identifying as Asian (10\% of workers vs. 6\% in the U.S. Census).

Our workers also tend to be more educated with 61\% reporting that they hold a Bachelor’s degree or higher, while the U.S. Census estimates 32\% of the general population holds a Bachelor’s degree or higher \cite{us-census2021}. Our workers are almost all employed either full or part-time (91\%), and largely live in urban or suburban areas (85\%). Geographic distribution across the U.S. shows more (35\%) workers in the South.

\section{Baselines} \label{baselines}

\begin{table}[h!]
\begin{center}
\begin{tabular}{l r}
\toprule
Model & Hits@1\\
\midrule
\textbf{\textsc{Language Models}}\\
\midrule
T5 & 0.28\\
T5 for CBQA (zero-shot) & 0.20\\
T5 for CBQA (fine-tuned) & \textbf{0.38} \\
\midrule
\textbf{\textsc{KGQA Models}}\\
\midrule
KVMemNet & 0.12\\
EmbedKGQA & 0.18\\
Rigel & 0.20\\
\midrule
\textbf{\textsc{Retriever-Reader Models}}\\
\midrule
DPR (zero-shot) & 0.15\\
DPR (trained) & 0.31\\
\bottomrule
\end{tabular}
\end{center}
\caption{Results of English baseline models on Mintaka}
\label{tab:baselines}
\end{table}

\begin{table*}[ht!]
\begin{center}
\begin{tabular}{l c c c c c c c c c}
\toprule
Model & multi & ar & de & es & fr & hi & it & ja & pt\\
\midrule
MT5 & 0.16 &  0.15 & 0.16 & 0.16 & 0.16 & 0.16 & 0.16 & 0.15 & 0.16\\
T5 for CBQA (translated) & 0.31 & 0.27 & 0.34 & 0.32 & 0.33 & 0.30 & 0.32 & 0.28 & 0.31 \\
Rigel & 0.19 & 0.18 & 0.19 & 0.19 & 0.19 & 0.17 & 0.19 & 0.20 & 0.18\\
\bottomrule
\end{tabular}
\end{center}
\caption{Results of baselines evaluated multilingually and in individual languages. Scores are reported as hits@1.}
\label{tab:multilingual-baselines}
\end{table*}

\subsection{Models}

We evaluate eight baselines on Mintaka. Since Mintaka contains only question and answer pairs, we only use models that can be trained end-to-end. We evaluate 3 language models, 3 knowledge graph-based models, and 2 retriever-reader models. For language and retriever-reader models, we use the answers written by the crowd workers in the Question Elicitation task as the label. For our knowledge graph, we use a Wikidata snapshot from October 18, 2021. We evaluate all of our baselines in English (Table \ref{tab:baselines}) and three of our baselines that could easily be set up multilingually in all languages (Table \ref{tab:multilingual-baselines}). Details on training data size can be found in Appendix \ref{appendix:b}.

\textbf{\textsc{T5 and MT5}} \cite{2020t5,xue-etal-2021-mt5} are baselines that only use a language model to predict answers to questions. We use the XL versions of T5 for English and MT5 for all other languages. We fine-tune both for 10,000 steps.

\textbf{\textsc{T5 for Closed Book QA (CBQA)}} \cite{2020t5cqba} is an extension of T5 that is fine-tuned as a QA model that can implicitly store and retrieve knowledge without an external source. We use \newcite{2020t5cqba}'s T5-XL model and evaluate on Mintaka both as zero-shot with a model fine-tuned on Natural Questions and with a model fine-tuned on Mintaka for 10,000 steps. We run an additional translation baseline where we automatically translate non-English questions to English using the M2M\_100 model \cite{fan2020beyond} and use our English model to return answers.
 
\textbf{\textsc{KVMemNet}}: Key-Value Memory Networks \cite{miller-etal-2016-key} work by first storing knowledge graph triples in a key-value structured memory. Then given a question, the model learns which keys are relevant to the question, and uses the values of those keys to return an answer. We follow the implementation by \newcite{shi2020kqa}.

\textbf{\textsc{EmbedKGQA}} \cite{saxena-etal-2020-improving} is a method that incorporates pre-trained knowledge graph embeddings into a KGQA model. EmbedKGQA consists of 1) a KG embedding module, 2) a question embedding module, and 3) an answer scoring module, which combines the question and KG embeddings to score and select answer entities. Since EmbedKGQA makes predictions over answer entities, we exclude questions where the answer is not an entity during training time, and count these as failures during test time. Scores for the subset of the test set that is answerable by the model can be found in Appendix \ref{appendix:b}.

\textbf{\textsc{Rigel}} \cite{oliya-etal-2021-end,sen-etal-2021-expanding} is an end-to-end question answering model based on ReifiedKB \cite{cohen2020scalable}. Rigel uses an encoder to encode the question and a decoder to return a probability distribution over all relations in the knowledge graph. The relations are followed in the knowledge graph to return predicted answers. For the encoder, we use RoBERTa \cite{liu2019roberta} for English and XLM-RoBERTa \cite{conneau-etal-2020-unsupervised} for all other languages. Again, since Rigel predicts over answer entities, we exclude questions where the answer is not an entity during training time and count them as failures during test time. 

\textbf{\textsc{Dense Passage Retrieval (DPR)}} \cite{karpukhin-etal-2020-dense} is a retriever-reader method that uses a dense retriever model to identify relevant Wikipedia passages given a question, followed by a reader model to score answer spans from the retrieved passages. For the retriever, we use \newcite{karpukhin-etal-2020-dense}'s model trained on Natural Questions, and for the reader, we evaluate both zero-shot with a model trained on Natural Questions and with a model trained on Mintaka. The reader sees the top 50 retrieved passages, and we take the highest-scoring span as the answer. 

\subsection{Analysis}

The results of the baselines in English in terms of hits@1 can be seen in Table \ref{tab:baselines}. Hits@1 is calculated based on the number of samples where the top prediction from the model matches the labeled answer, either as exact string match for text answers or as entity IDs for entity answers. A further breakdown per complexity type is in Appendix \ref{appendix:c}. The best-performing model is the fine-tuned T5 for Closed Book QA with 38\% hits@1. An analysis of the outputs shows that even though the model does not have access to an external knowledge source, it does recall factual information, such as the capital of Iraq is Baghdad. For more complex questions, the model can usually predict in the correct neighborhood of the answer. For example, for "\textit{What is the second Marvel Cinematic Universe movie chronologically?}", the model predicts "\textit{Thor: The Dark World}", which is a Marvel movie, however it lacks the complex reasoning functionality to calculate the second movie chronologically.

The trained DPR model has the second highest score with 31\% hits@1. We find that DPR can handle complex questions, as long as the complex reasoning is already done in the passage. For example, the model can answer "\textit{When did Roger Federer win his first Grand Slam?}" with "\textit{2003}" from a passage that includes, "Roger Federer won his first Grand Slam title in the 2003 Wimbledon Championships." However in cases where the reasoning is not included in the passage, the model struggles. For example given the same question about the second Marvel movie, the model predicts Iron Man and fails to find a passage explicitly mentioning the second movie chronologically. 

Finally, the best-performing KGQA model is Rigel with 20\% hits@1. Our KGQA baselines handle only entity answers and can only traverse the knowledge graph by following relations, so they are limited. Although the KGQA models score lower than the other models, they do have advantages. Language models at the scale of T5-XL are computationally expensive and the knowledge stored in the parameters is static. Knowledge graph based methods like Rigel can be updated easily by updating the external knowledge graph and can also return more interpretable answers. Given the Marvel question again, Rigel predicts a path from the Marvel Cinematic Universe to all the Marvel films. Rigel can't perform sorting or filtering, but it's possible to see what the next steps should be: identifying the chronological order, ordering by chronological order, and finding the second in the ordered list. Understanding how a model has arrived at an answer and what steps should be added to arrive at the correct answer are useful features for debugging and improving KGQA models.

Table \ref{tab:multilingual-baselines} shows results on Mintaka evaluated both multilingually on all languages and in each language individually. For almost all models, the results are slightly lower than in English. The MT5 language model has worse performance than the English T5 model, which may be because unlike T5, MT5 is not pre-trained on any supervised tasks. Our T5 for CBQA model using English translations outperforms MT5, however the scores are still lower than on the original English questions, so the automatic translations do degrade performance. For Rigel, the main gap to English is on the encoding side, where we use XLM-RoBERTa instead of RoBERTa, showing a gap in the performance between the multilingual encoder and English encoder. All models show that there is still work needed for parity across all languages. 

Overall, the baselines show that Mintaka is a challenging dataset. None of our baselines explicitly handle all of the complexity types available in Mintaka. The language model-only models especially struggle to handle questions that require numerical operations such as counts. The knowledge graph-based models rely on relation following to traverse the knowledge graph to an answer. This prevents models from correctly predicting answers that require more complex operations, even if the facts required are available in the knowledge graph. Adding additional operations and learning to select the correct operation for each question could lead to significant improvement. A combination of powerful language models, potentially to encode questions or identify question entities, with the interpretable facts and operations available in a knowledge graph is a promising direction to create better models on Mintaka. 

\section{Conclusions} \label{conclusions}
In this paper, we introduce Mintaka, an end-to-end question answering dataset linked to Wikidata. Mintaka addresses an important gap in QA datasets by being large-scale, complex, naturally-elicited, and multilingual. Our baselines show that there is room for improvement in existing methods to handle complex questions, especially in all languages. With the release of Mintaka, we hope to encourage researchers to continue pushing the boundaries of question answering to handle more complex questions in more languages.

\bibliography{anthology,custom}
\bibliographystyle{acl_natbib}

\appendix
\section{MTurk Tasks} \label{appendix:a}

\begin{figure*}[h!]
    \includegraphics[width=\textwidth]{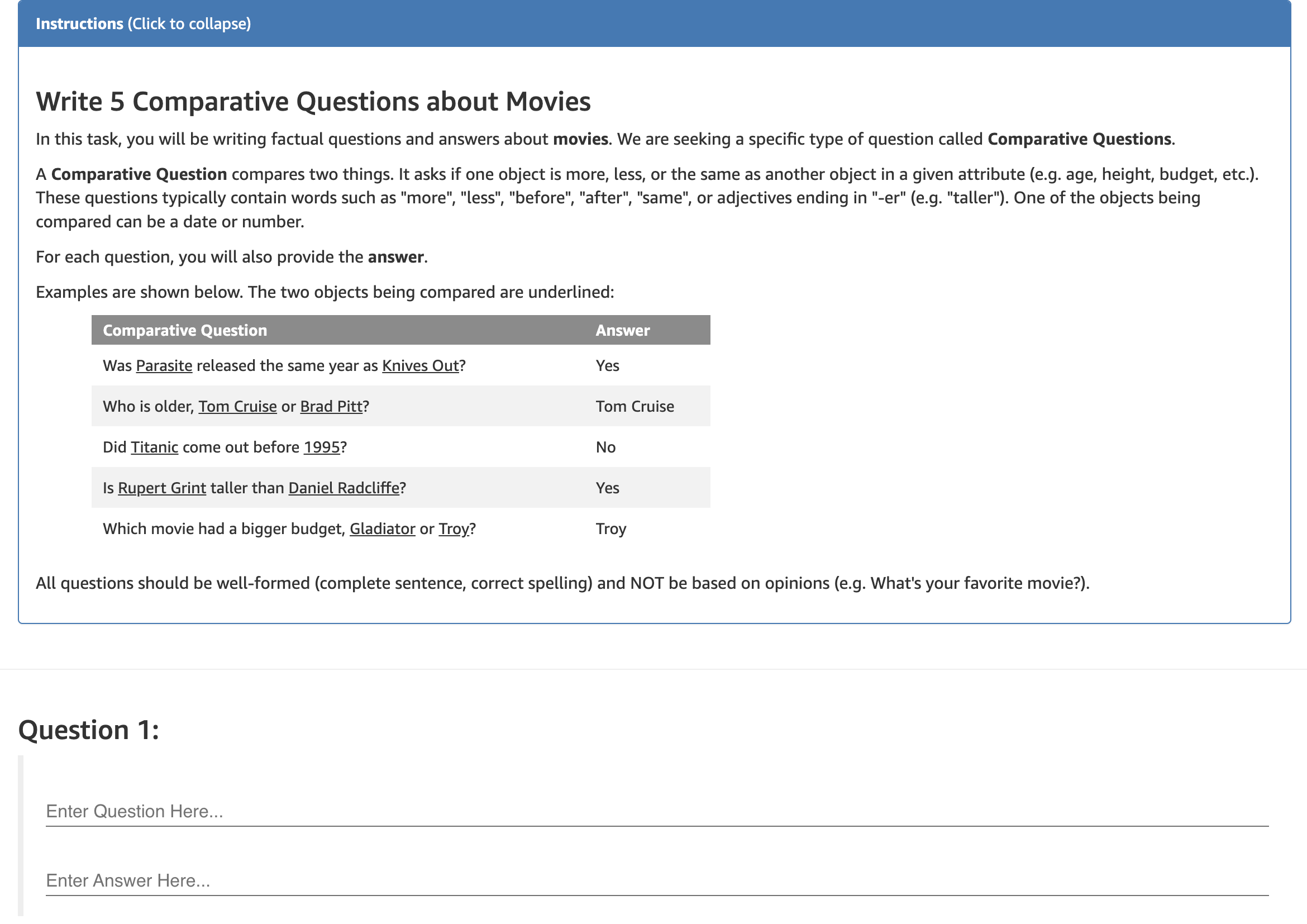}
    \caption{An example of the question elicitation MTurk task, where a worker is asked to write comparative questions about movies}
\label{fig:question-elicit}
\end{figure*}

\begin{figure*}[h!]
    \includegraphics[width=\textwidth]{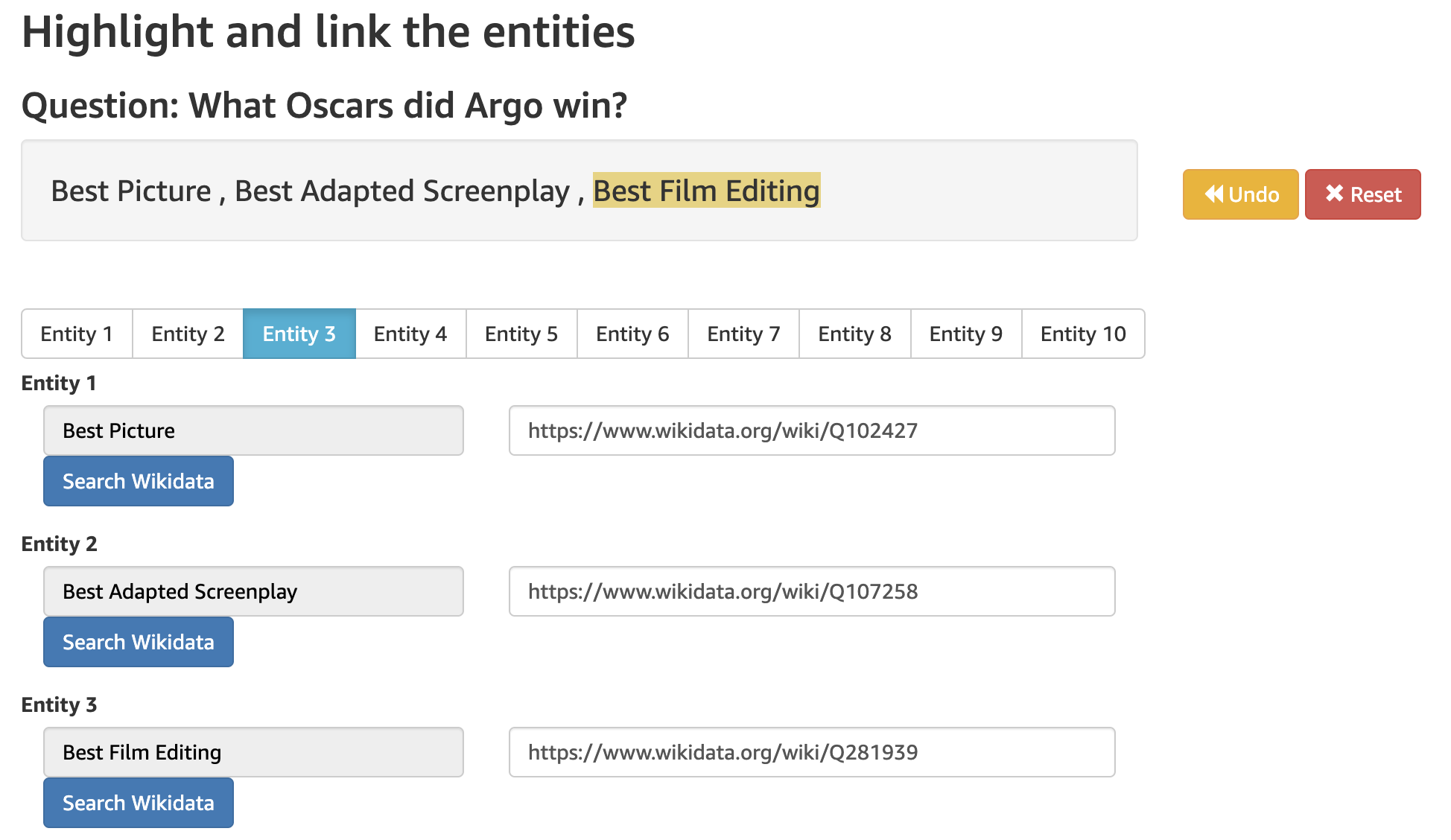}
    \caption{An example of the answer entity annotation MTurk task, where a worker is asked to identify and link the entities in the answer "Best Picture, Best Adapted Screenplay, Best Film Editing".}
\label{fig:answer-ent}
\end{figure*}

\begin{figure*}[h!]
    \includegraphics[width=\textwidth]{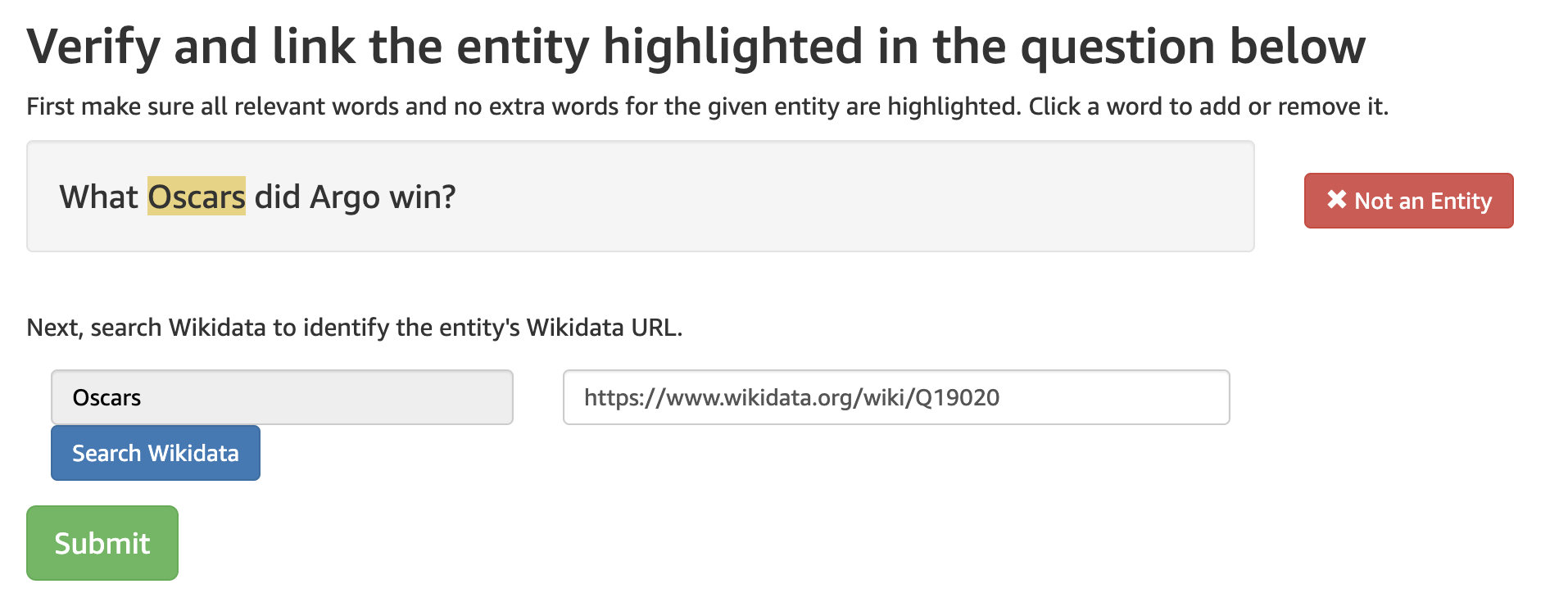}
    \caption{Example of the question entity annotation MTurk task, where the entity is already highlighted (in this case, "Oscars"), and the worker is asked to verify or modify the highlighted string and then link to Wikidata.}
\label{fig:question-ent}
\end{figure*}

Figures \ref{fig:question-elicit}, \ref{fig:answer-ent}, and \ref{fig:question-ent} show the interfaces used by MTurk workers to complete each of the tasks to build the Mintaka dataset. All of these tasks were hosted on MTurk. Figure \ref{fig:question-elicit} is the Question Elicitation task. This example is for writing questions about the topic \textsc{Movies} and the complexity type \textsc{Comparative}. In each task, a worker would be shown examples and asked to write five questions. 

Figure \ref{fig:answer-ent} is the Answer Entity Annotation task. In this example, a worker is shown the question-answer pair "Q: What Oscars did Argo win? A: Best Picture, Best Adapted Screenplay, Best Film Editing" and asked to identify the entities in the answer. The question is given as context, allowing the worker to know that these awards refer to Oscar awards. After highlighting each entity, the "Search Wikidata" button is automatically populated to create a search link on Wikidata for the given string that will open in a new window or tab. The workers could then look at all the choices and enter the URL of the correct entity.

Figure \ref{fig:question-ent} is the Question Entity Annotation task. This example again shows the question "What Oscars did Argo win?" with "Oscars" highlighted. Workers were asked to focus on one entity at a time, so even though "Argo" is also a valid entity in this question, for this task, we are only interested in linking "Oscars". Workers would first verify that Oscars is a valid and complete entity, or modify the string if there was an error. Then, similar to the Answer Entity task, the "Search Wikidata" button lets worker search the string on Wikidata and find the URL of the correct entity.

\section{Model Training Details} \label{appendix:b}

Table \ref{tab:training-details} shows the train, dev, and test set sizes of each of the models. The Hits@1 Subset score is the score based on the test subset answerable by the model, and the Hits@1 Adjusted score is the adjusted hits@1 to account for the full test set. The full Mintaka dataset has 14,000 train examples, 2,000 dev examples, and 4,000 test examples.  

For T5, fine-tuned T5 for CBQA, KVMemNet, and trained DPR we use the full train and dev sets for training and the full test set for evaluation. For MT5, we fine-tune on all languages simultaneously and report results overall and for each language individually. We found that fine-tuning MT5 on individual languages returned similar scores but at a higher computational cost. The zero-shot T5 CBQA and DPR models have no train or dev set since we evaluate them directly on the test set. The EmbedKGQA and Rigel models only predict over entities in a knowledge graph. This means that any sample that doesn't have at least one entity in the question and one entity in the answer is not used in the training or dev sets. This excludes samples where the answer is a number, a boolean (all yes/no questions), a date, or a string, or where an entity was found but no Wikidata link existed (for example, if the name of a video game character was identified, but no Wikidata ID existed). EmbedKGQA loses some additional examples if an entity was found in Wikidata but didn't have entity-to-entity facts, which are used to build the KG embeddings. For example, some entities in Wikidata only exist with labels (entity-to-string) facts but lack entity-to-entity facts.

\section{Mintaka Examples} \label{appendix:c}

Table \ref{tab:mintaka-examples} shows additional examples of question-answer pairs from the Mintaka dataset.

\begin{table*}
\begin{tabular}{l}
\toprule
\textbf{Q:} Which series is older, Metroid (Q12397) or Super Mario Bros (Q23902998)?\\
\textbf{A:} Super Mario Bros (Q23902998)\\
\midrule
\textbf{Q:} What year was the first (number: 1) book of the A Song of Ice and Fire (Q45875) series published?\\
\textbf{A:} 1996\\
\midrule
\textbf{Q:} Which Amon Amarth (Q192863) albums did Fredrik Andersson (Q3752814) not perform as the drummer? \\
\textbf{A:} Once Sent from the Golden Hall (Q1366410), Jomsviking (Q22674162), Berserker (Q62272261)\\
\midrule
\textbf{Q:} Is the Eiffel Tower (Q243) located in Italy (Q38)?\\
\textbf{A:} No \\
\midrule
\textbf{Q:} Who was the president of Argentina (Q414) from 1989 (date: 1989) to 1999 (date: 1999)?\\
\textbf{A:} Carlos Menem (Q185107)\\
\midrule
\textbf{Q:} How many teams has Matthew Stafford (Q889130) played for?\\
\textbf{A:} 2: Detroit Lions (Q271880) and Los Angeles Rams (Q337377)\\
\midrule
\textbf{Q:} What is the name of the star of Iron Man's (Q192724) wife?\\
\textbf{A:} Susan Downey (Q936542)\\
\midrule
\textbf{Q:} What is the third (number: 3) longest river in the USA (Q30)?\\
\textbf{A:} Yukon River (Q104437)\\
\midrule
\textbf{Q:} Who ruled for a longer period of time, King Tut (Q12154) or Alexander the Great (Q8409)?\\
\textbf{A:} Alexander the Great (Q8409)\\
\bottomrule
\end{tabular}
\caption{Example question-answer pairs from Mintaka. Question and answer annotations are shown here in-line with Wikidata Q-codes.}
\label{tab:mintaka-examples}
\end{table*}

\section{Model Training Details} \label{appendix:d}

\begin{table*}[h]
\begin{center}
\begin{tabular}{l l r r r r r r}
\toprule
& & & & & Hits@1 & Hits@1\\
Model & Lang & Train & Dev & Test & Subset & Adjusted\\
\midrule
T5 & en & 14,000 & 2,000 & 4,000 & 0.28 & 0.28\\
MT5 & ar & 126,000 & 18,000 & 36,000 & 0.15 & 0.15 \\
& de & 126,000 & 18,000 & 36,000 & 0.16 & 0.16 \\
& es & 126,000 & 18,000 & 36,000 & 0.16 & 0.16 \\
& fr & 126,000 & 18,000 & 36,000 & 0.16 & 0.16 \\
& hi & 126,000 & 18,000 & 36,000 & 0.16 & 0.16 \\
& it & 126,000 & 18,000 & 36,000 & 0.16 & 0.16 \\
& ja & 126,000 & 18,000 & 36,000 & 0.15 & 0.15\\
& pt & 126,000 & 18,000 & 36,000 & 0.16 & 0.16 \\
& multi & 126,000 & 18,000 & 36,000 & 0.16 & 0.16 \\
\midrule
T5 CBQA (zero-shot) & en & -- & -- & 4,000 & 0.20 & 0.20 \\
T5 CBQA (fine-tuned) & en & 14,000 & 2,000 & 4,000 & 0.38 & 0.38 \\
T5 CBQA (translated) & ar & 14,000 & 2,000 & 4,000 & 0.27 & 0.27 \\
& de & 14,000 & 2,000 & 4,000 & 0.34 & 0.34 \\
& es & 14,000 & 2,000 & 4,000 & 0.32 & 0.32 \\
& fr & 14,000 & 2,000 & 4,000 & 0.33 & 0.33 \\
& hi & 14,000 & 2,000 & 4,000 & 0.30 & 0.30 \\
& it & 14,000 & 2,000 & 4,000 & 0.32 & 0.32 \\
& ja & 14,000 & 2,000 & 4,000 & 0.28 & 0.28\\
& pt & 14,000 & 2,000 & 4,000 & 0.31 & 0.31 \\
& multi & 14,000 & 2,000 & 4,000 & 0.31 & 0.31 \\
\midrule
KVMemNet & en & 14,000 & 2,000 & 4,000 & 0.12 & 0.12 \\
\midrule
EmbedKGQA & en & 9,837 & 1,409 & 2,809 & 0.26 & 0.18\\
\midrule
Rigel & en & 9,839 & 1,409 & 2,809 & 0.29 & 0.20 \\
& ar & 9,839 & 1,409 & 2,809 & 0.26 & 0.18 \\
& de & 9,839 & 1,409 & 2,809 & 0.27 & 0.19 \\
& es & 9,839 & 1,409 & 2,809 & 0.27 & 0.19 \\
& fr & 9,839 & 1,409 & 2,809 & 0.27 & 0.19 \\
& hi & 9,839 & 1,409 & 2,809 & 0.25 & 0.17 \\
& it & 9,839 & 1,409 & 2,809 & 0.27 & 0.19 \\
& ja & 9,839 & 1,409 & 2,809 & 0.28 & 0.20\\
& pt & 9,839 & 1,409 & 2,809 & 0.25 & 0.18 \\
& multi & 88,551 & 12,681 & 25,281 & 0.27 & 0.19 \\
\midrule
DPR (zero-shot) & en & -- & -- & 4,000 & 0.15 & 0.15 \\
DPR (trained) & en & 14,000 & 2,000 & 4,000 & 0.31 & 0.31 \\
\bottomrule
\end{tabular}
\end{center}
\caption{Details of the train, dev, and test set sizes for all models. Hits@1 Subset shows the hits@1 score on the available test set. Hits@1 Adjusted adjusts the hits@1 score for the full test set of 4,000 questions.}
\label{tab:training-details}
\end{table*}

Table \ref{tab:complexity-breakdown} shows the breakdown of performance by complexity type for all trained models. For count questions, we allow models to return the entities that are being counted rather than the number. For example, if the question is "\textit{How many Academy Awards has Jake Gyllenhaal been nominated for?}", we allow the model to return "Academy Award for Best Supporting Actor" rather than "1". For entity answers, the order of answers does not matter, but for text answers, we use exact string matching.

The results show that both complex and generic questions remain challenging for models. For generic questions, this shows that even though we didn't specify a complexity type, these questions are not trivial. On complex questions, some of our models do perform better on comparative and yes or no questions. However for these questions, there is usually either a choice between two entities or a choice between "Yes" or "No", so randomly guessing would score 50\%. This means that models scoring around 50\% are not necessarily performing the reasoning required.

\begin{table*}[h]
\begin{center}
\begin{tabular}{l l r r r r r r r r r}
\toprule
Model & Lang & Gen & Mhop & Intsct & Diff & Comp & Superl & Ord & Count & YesNo\\
\midrule
T5 & en & 0.24 & 0.14 & 0.32 & 0.13 & 0.55 & 0.31 & 0.14 & 0.05 & 0.67\\
MT5 & multi & 0.06 & 0.05 & 0.12 & 0.06 & 0.45 & 0.13 & 0.04 & 0.01 & 0.58
\\
& ar & 0.05 & 0.04 & 0.11 & 0.06 & 0.45 & 0.14 & 0.03 & 0.01 & 0.57
\\
& de & 0.07 & 0.05 & 0.13 & 0.07 & 0.42 & 0.14 & 0.05 & 0.01 & 0.57
\\
& es & 0.07 & 0.06 & 0.13 & 0.07 & 0.48 & 0.12 & 0.05 & 0.02 & 0.57
\\
& fr & 0.07 & 0.06 & 0.13 & 0.06 & 0.46 & 0.14 & 0.05 & 0.01 & 0.57
\\
& hi & 0.04 & 0.05 & 0.11 & 0.06 & 0.48 & 0.14 & 0.05 & 0.01 & 0.59
\\
& it & 0.07 & 0.06 & 0.12 & 0.07 & 0.45 & 0.14 & 0.05 & 0.01 & 0.58
 \\
& ja & 0.05 & 0.04 & 0.10 & 0.04 & 0.45 & 0.13 & 0.03 & 0.01 & 0.59
\\
& pt & 0.07 & 0.05 & 0.14 & 0.07 & 0.45 & 0.13 & 0.05 & 0.01 & 0.59
\\
\midrule
T5 CBQA \\
\hspace{5pt} zero-shot & en & 0.31 & 0.15 & 0.31 & 0.13 & 0.25 & 0.28 & 0.22 & 0.00 & 0.03\\
\hspace{5pt} fine-tuned & en & 0.41 & 0.21 & 0.44 & 0.21 & 0.58 & 0.42 & 0.30 & 0.09 & 0.71\\
\hspace{5pt} translated & multi & 0.32 & 0.15 & 0.38 & 0.16 & 0.51 & 0.36 & 0.26 & 0.07 & 0.56\\
& ar & 0.23 & 0.12 & 0.31 & 0.14 & 0.49 & 0.31 & 0.20 & 0.05 & 0.58\\
& de & 0.36 & 0.17 & 0.44 & 0.17 & 0.59 & 0.37 & 0.27 & 0.07 & 0.62\\
& es & 0.35 & 0.15 & 0.40 & 0.19 & 0.54 & 0.38 & 0.27 & 0.08 & 0.55\\
& fr & 0.35 & 0.17 & 0.42 & 0.17 & 0.52 & 0.38 & 0.28 & 0.07 & 0.60\\
& hi & 0.30 & 0.14 & 0.34 & 0.16 & 0.49 & 0.34 & 0.25 & 0.05 & 0.65\\
& it & 0.35 & 0.19 & 0.42 & 0.17 & 0.50 & 0.38 & 0.29 & 0.08 & 0.49\\
& ja & 0.29 & 0.13 & 0.33 & 0.13 & 0.45 & 0.36 & 0.25 & 0.06 & 0.55\\
& pt & 0.36 & 0.17 & 0.41 & 0.16 & 0.51 & 0.39 & 0.28 & 0.07 & 0.44\\
\midrule
KVMemNet & en & 0.04 & 0.03 & 0.06 & 0.04 & 0.29 & 0.12 & 0.06 & 0.01 & 0.52\\
\midrule
EmbedKGQA & en & 0.15 & 0.07 & 0.18 & 0.18 & 0.05 & 0.40 & 0.18 & 0.44 & 0.00\\
\midrule
Rigel & en & 0.19 & 0.09 & 0.30 & 0.13 & 0.43 & 0.13 & 0.09 & 0.50 & 0.00
\\
& multi & 0.17 & 0.09 & 0.25 & 0.12 & 0.45 & 0.11 & 0.08 & 0.44 & 0.00\\
& ar & 0.17 & 0.09 & 0.25 & 0.12 & 0.39 & 0.09 & 0.08 & 0.45 & 0.00 \\
& de & 0.18 & 0.09 & 0.25 & 0.12 & 0.45 & 0.10 & 0.09 & 0.46 & 0.00 \\
& es & 0.18 & 0.09 & 0.26 & 0.12 & 0.47 & 0.11 & 0.08 & 0.47 & 0.00\\
& fr & 0.18 & 0.10 & 0.27 & 0.13 & 0.43 & 0.11 & 0.08 & 0.45 & 0.00\\
& hi & 0.15 & 0.08 & 0.24 & 0.11 & 0.42 & 0.08 & 0.08 & 0.42 & 0.00\\
& it & 0.17 & 0.08 & 0.26 & 0.12 & 0.46 & 0.11 & 0.07 & 0.47 & 0.00\\
& ja & 0.19 & 0.09 & 0.28 & 0.12 & 0.45 & 0.10 & 0.08 & 0.46 & 0.00\\
& pt & 0.15 & 0.07 & 0.24 & 0.10 & 0.44 & 0.09 & 0.08 & 0.44 & 0.00\\
\midrule
DPR \\
\hspace{5pt} zero-shot & en & 0.27 & 0.10 & 0.20 & 0.10 & 0.12 & 0.23 & 0.21 & 0.03 & 0.00\\
\hspace{5pt} trained & en & 0.31 & 0.14 & 0.37 & 0.19 & 0.47 & 0.29 & 0.27 & 0.40 & 0.39\\
\bottomrule
\end{tabular}
\end{center}
\caption{A breakdown by complexity type for all trained baselines. Some complexity types are abbreviated: "Gen" is Generic questions, "Mhop" is multihop questions, "Intsct" is Intersection questions, "Diff" is Difference questions, "Comp" is Comparative questions, "Superl" is Superlative questions, and "Ord" is ordinal questions.}
\label{tab:complexity-breakdown}
\end{table*}

\end{document}